\documentclass{article}

\PassOptionsToPackage{square,numbers}{natbib}

     \usepackage[final]{neurips_2019}

\usepackage[utf8]{inputenc} % allow utf-8 input
\usepackage[T1]{fontenc}    % use 8-bit T1 fonts
\usepackage{hyperref}       % hyperlinks
\usepackage{url}            % simple URL typesetting
\usepackage{booktabs}       % professional-quality tables
\usepackage{amsfonts}       % blackboard math symbols
\usepackage{nicefrac}       % compact symbols for 1/2, etc.
\usepackage{microtype}      % microtypography
\usepackage{graphicx}
\usepackage{caption}
\usepackage{subcaption}
\usepackage{amsmath}
\usepackage{float}

\usepackage{todonotes}

\DeclareMathOperator*{\argmin}{\arg\!\min}

\title{Learning Visual Dynamics Models of Rigid Objects using Relational Inductive Biases}

\author{%
  Fabio Ferreira$^{1,2}$ \\
  \texttt{fabiof@cs.stanford.edu} \\
  \And
  Lin Shao$^{2}$ \\
  \And
  Tamim Asfour$^1$
  \And
  Jeannette Bohg$^{2}$ \vspace{-2ex}\\
  \And
  \normalfont $^1$Institute for Anthropomatics and Robotics, Karlsruhe Institute of Technology (KIT)\\
  $^2$Computer Science Department, Stanford University
}

\begin{document}

\maketitle

\begin{abstract}
Endowing robots with human-like physical reasoning abilities remains challenging. We argue that existing methods often disregard spatio-temporal relations and by using \emph{Graph Neural Networks} (GNNs) that incorporate a relational inductive bias, we can shift the learning process towards exploiting relations. In this work, we learn action-conditional forward dynamics models of a simulated manipulation task from visual observations involving cluttered and irregularly shaped objects. We investigate two GNN approaches and empirically assess their capability to generalize to scenarios with novel and an increasing number of objects. The first, \emph{Graph Networks} (GN) based approach, considers explicitly defined edge attributes and not only does it consistently underperform an auto-encoder baseline that we modified to predict future states, our results indicate how different edge attributes can significantly influence the predictions. Consequently, we develop the \emph{Auto-Predictor} that does not rely on explicitly defined edge attributes. It outperforms the baseline and the GN-based models. Overall, our results show the sensitivity of GNN-based approaches to the task representation, the efficacy of relational inductive biases and advocate choosing lightweight approaches that implicitly reason about relations over ones that leave these decisions to human designers.
\end{abstract}

\section{Introduction}
Physical reasoning from visual input is a central aspect of human intelligence \cite{Spelke2006}. However, endowing robots with this ability still remains challenging. We hypothesize that exploiting spatio-temporal relations is beneficial for physical reasoning.
To test this hypothesis, we use Graph Neural Networks (GNNs) \cite{Scarselli2009} that incorporate a relational inductive bias. This work seeks to understand whether GNNs are beneficial within a model-based policy for vision-based, robotic manipulation.

A popular approach for modeling visual dynamics is based on video prediction methods that treat dynamics prediction as an image-to-image translation \cite{Finn2016, finn2017deep, agrawal2016learning, villegas2017learning, wichers2018hierarchical}. These methods often disregard the structure in the input, i.e. ignore that the input consists of entities that relate to each other in specific ways. Hence, a different strategy is to exploit this structure by relying on object-centric or graph representations. For example, \cite{Watters2017} process visual input for learning a physics model of 2D objects to predict future object-centric states. Despite accurate predictions, the visual and object dynamics complexity is low in the considered scenarios and predictions are not conditioned on actions. However, conditioning a prediction on a potential action is essential when planning an optimal sequence of actions towards a given goal. \cite{ye2019compositional} develop an entity-centric dynamics model with a probabilistic prediction module that uses region proposals of 3D entities. While the approach shows great prediction performance on visually complex scenarios, the predictions are also not conditioned on actions. \cite{Watters2019} learn a dynamics model of 2D objects conditioned on actions. However, since the scenario complexity is low (e.g. no object collisions), it is unclear whether the proposed method applies to scenarios with high visual and dynamic complexity. \cite{Janner2019} learn an action-conditional dynamics model for a 3D object stacking task to predict a steady-state configuration of an object after applying an action. Although \cite{Janner2019} train their predictive model on a scenario with a high visual complexity, the action space is comprised of pre-defined, abstract primitives rather than continuous control inputs as in our work. This makes it difficult to apply the method to tasks where a long sequence of continuous actions is required to achieve a goal.

This work addresses the aforementioned limitations by learning dynamics models with GNNs. These GNNs predict visual future states of a simulated, visually complex object manipulation task conditioned on low-level actions. We investigate two approaches and compare their predictive capability in experiments over three datasets. The first approach is based on the \emph{Graph Networks} (GN) framework \cite{Battaglia2018} and considers explicitly defined edge attributes. Our results indicate how different types of edge attributes can affect the performance of the GN-based models and that a simple auto-encoder baseline consistently outperforms all these models. In light of these results, we develop our second approach, the \emph{Auto-Predictor} (AP), which is similar to the \emph{O2P2} architecture in \cite{Janner2019}. It extends the auto-encoder baseline by GNN functions for computing object state updates and object interactions without relying on explicitly defined edge attributes. This results in a lightweight approach that outperforms both our baseline and the GN-based models in the majority of our experiments. This work presents the insights we gained from our study on a task that is complex both visually and in terms of object dynamics. In the context of learning dynamics models with GNNs, the results show that 1) sub-optimal selections of edge attributes can significantly impact the predictive performance and 2) lightweight GNN approaches which implicitly reason about relations can be superior to approaches that require explicitly-defined relation attributes.\footnote{\url{https://sites.google.com/view/dynamicsmodels/}}

\section{Method}
To assess a GNN-based model's prediction capabilities, we assume access to the segmentation of a scene image where an object state is represented by the corresponding object segmentation mask and the full-scene RGB and depth frame. In our approaches, the prediction of an object's state is based on its current state, the states of other objects and a future control (action) input. For this, we train and test our models on data that is collected from a simulated \emph{object singulation} task in which a robot tries to physically separate a set of cluttered objects through manipulation \cite{Eitel2017}. Our task scenario consists of at least three and up to five asymmetrically-shaped, rigid 3D objects. A robot arm equipped with a gripper executes pushing actions to singulate the objects placed inside a container. Each episode consists of $T \in [7,...,50]$ total time steps and at each time step $t$ the simulation provides visual data from an angled top perspective, i.e. an RGB frame $\mathcal{I}^{(t)}$, a depth frame $\mathcal{D}^{(t)}$ and a segmentation frame $\mathcal{S}^{(t)}$. The segmentation frame $\mathcal{S}^{(t)}$ is composed of several individual object segmentation masks. For evaluating the predictive performance of our models under varying data availability, we also access unobservable states in some of our experiments, i.e. object centroid position $\mathcal{O}_{pos}^{(t)} \in \mathbb{R}^3$ and linear velocity $\mathcal{O}_{vel}^{(t)} \in \mathbb{R}^3$. We also know the gripper position and gripper linear velocity at the next time step $t+1$ which we concatenate to a future control input and denote as $\mathcal{C}^{(t)}$. In total, we generated three datasets: \emph{3 objects}, \emph{5 objects (2 novel)}, \emph{5 objects (5 novel)}. The first is used for training only. The remaining are used for testing only and add two novel objects to \emph{3 objects} or replace all objects by five novel objects, respectively. Each dataset episode with time steps $t \in [1, ..., T]$ is transformed into a sequence of attributed graphs $(G^{(1)}, G^{(2)}, ..., G^{(T))})$ where the data available at time step $t$ is assigned to graph $G^{(t)}$. We represent each object at time $t$ as a node by $v^{(t)} = (\mathcal{I}^{(t)}, \mathcal{S}_{v, mask}^{(t)}, \mathcal{D}^{(t)}, \mathcal{O}_{v, pos}^{(t)}, \mathcal{O}_{v, vel}^{(t)})$ and connect all nodes with all other nodes through directed edges $e^{(t)} = (\mathcal{O}_{v, vel}^{(t)}, \mathcal{O}_{v, pos}^{(t-1)}, \mathcal{O}_{v, pos}^{(t)})$ for a sending node $v$. The idea behind using object velocity and position for the edges is to allow the nodes to reason about object proximity and the effects of collisions caused by other objects. $\mathcal{S}_{v, mask}^{(t)}$ is hereby a frame in which all pixels are set to zero except the pixels that correspond to the object, represented by node $v$, are set to one. It should be noted that if $t=1$, we set the previous object position $\mathcal{O}_{v, pos}^{(t-1)} = \mathcal{O}_{v, pos}^{(t)}$. 

\begin{figure}[ht!]
\centering
\begin{subfigure}[b]{0.45\textwidth}
    \centering
    \includegraphics[width=0.9\linewidth]{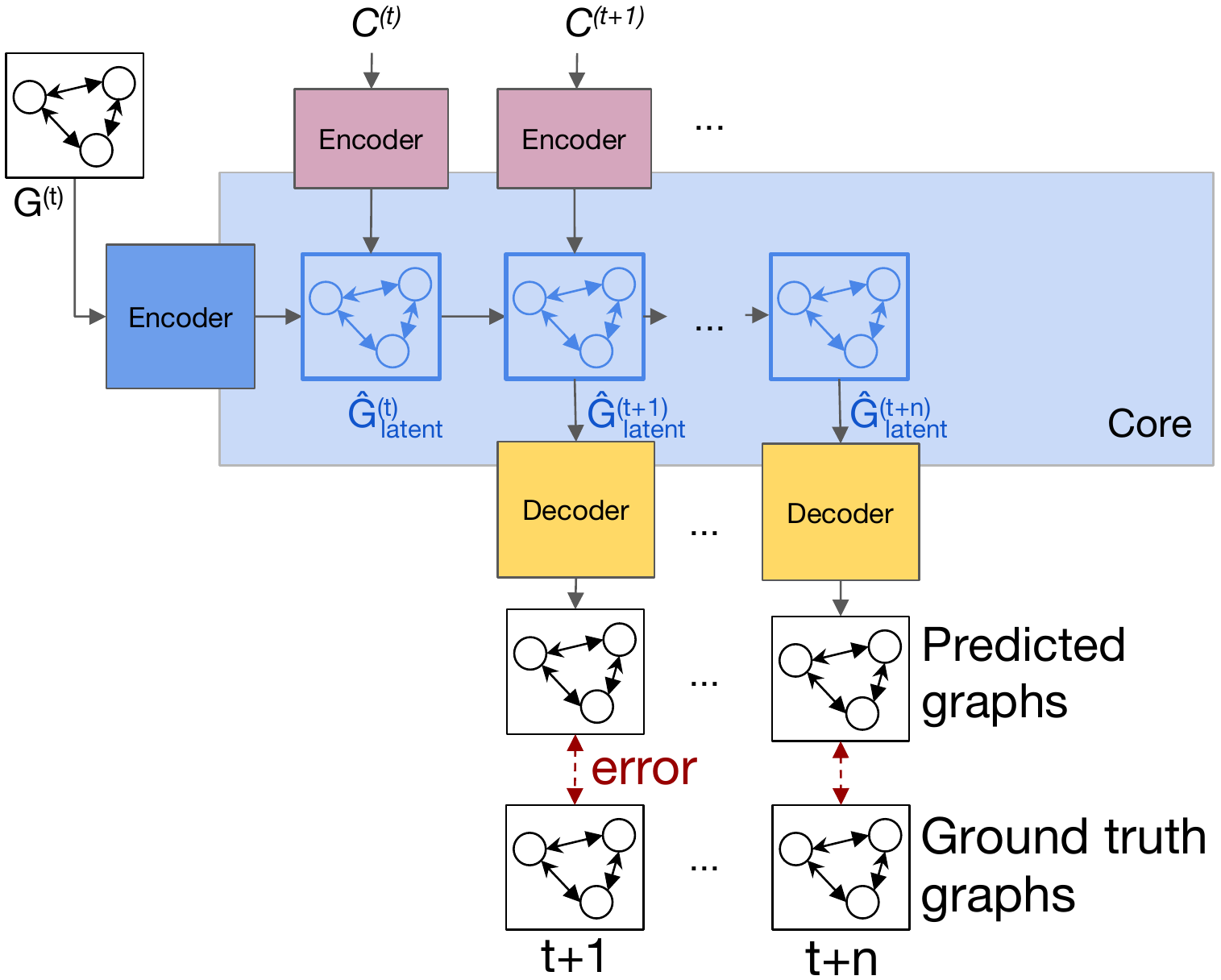}
    \caption{}
    \label{fig:gnmodel}
\end{subfigure}%
\begin{subfigure}[b]{0.45\textwidth}
    \centering
    \raisebox{5mm}{
    \includegraphics[width=0.9\linewidth]{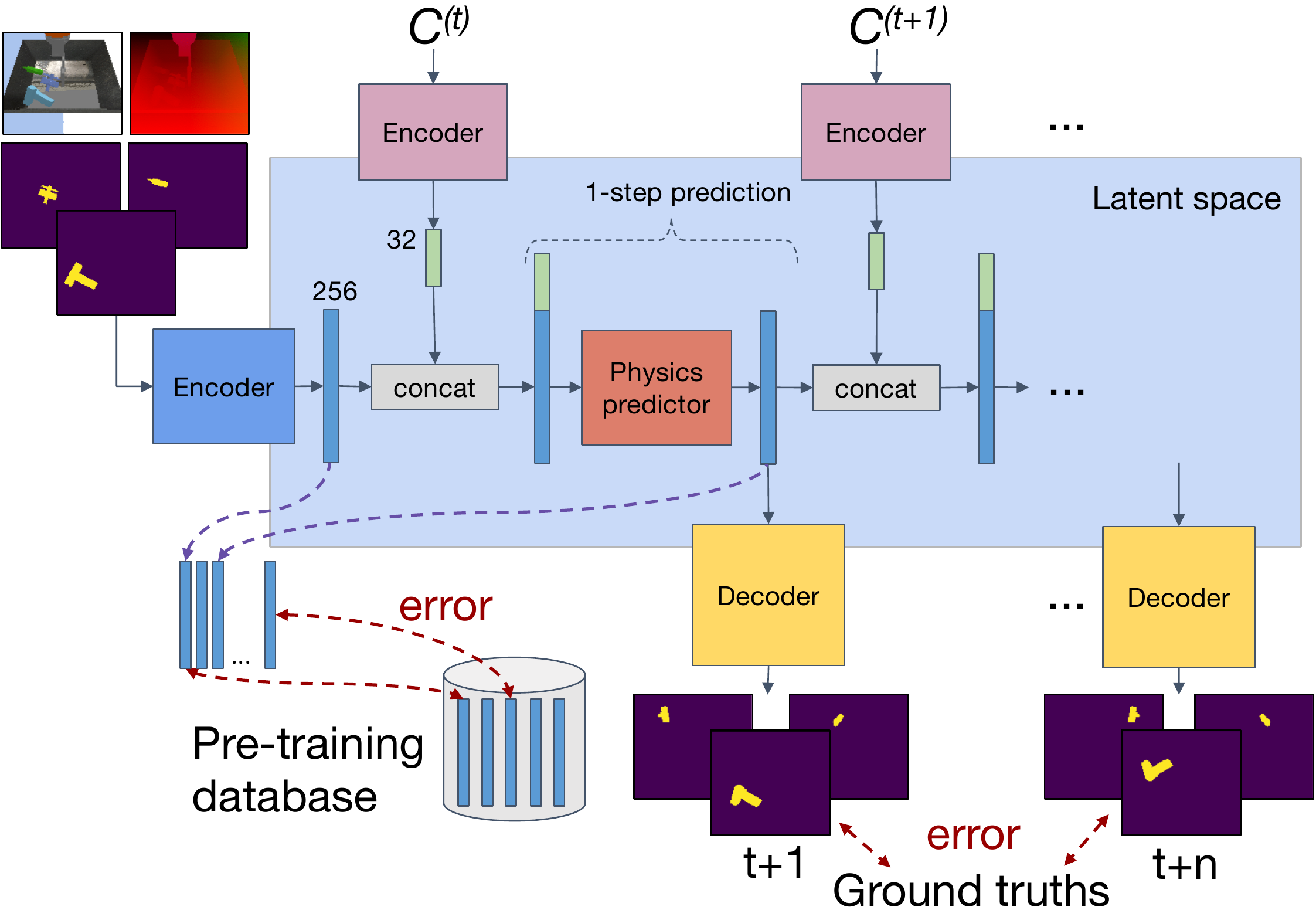}}
    \caption{}
    \label{fig:apmodel}
\end{subfigure}
\caption{a) The GN architecture b) The AP architecture (we depict the vectors for one object only)}
\end{figure}

\paragraph{Graph Networks Approach}
For the GN approach, we adopt the \emph{Encode-Process-Decode} architecture \cite{Battaglia2018} (see Figure \ref{fig:gnmodel}). The GN framework uses an extended definition of a graph $(u,V,E)$ where $u$ denotes the global attribute that operates on graph-level (e.g. $u$ can represent gravity) and which we use for our control inputs. The architecture first \emph{encodes} each element of an input graph $G^{(t)}$ into a latent graph $\hat{G}^{(t)}_{latent}$ with an independent recurrent block. Then, it updates each element of $\hat{G}^{(t)}_{latent}$ in the \emph{processing phase} with the full GN block $n$ times, yielding $n$ latent graphs. Lastly, it \emph{decodes} the resulting individual elements of these graphs back into the original space with an independent recurrent block to get $(\hat{G}^{(t+1)}, ..., \hat{G}^{(t+n)})$. Since our task requires the control input $\mathcal{C}^{(t)}$ to be re-set at every step, we replace the $u$ attribute of each latent graph with a new encoded control input  $\hat{G}^{(t)}_{latent} = (\hat{G}^{(t)}_{latent} \setminus \{u\}) \cup \{\mathcal{C}^{(t)}_{latent}\}$. We use a CNN encoder and a transpose CNN decoder for the nodes and Multilayer Perceptrons (MLPs) for encoding the control input, for encoding/decoding edge and node (position and velocity) attributes and for functions in the full GN block (see supplementary materials for more details). By using the data in nodes and edges as described above, we optimize the GN to predict future edge $\hat{e}_i$ and node states of an $i$-th node, where the prediction of a nodes state is composed of two individual predictions: first, the visual state $\hat{\mathcal{S}}_{i, mask}$ optimized by a binary cross entropy error $BCE(p,q) = -p \log q - (1-p)\log (1-q)$ and second, the non-visual state represented by position and velocity $\hat{p}_{i}$ which, together with the edge, is optimized with mean squared error terms as follows:
\begin{equation}
    \begin{aligned}
         \argmin \frac{1}{n} \sum_{t=1}^{n} \big(
         \frac{1}{N^e}\sum_{i=1}^{N^e} ( e_i^{(t)} - \hat{e_i}^{(t)} )^2
         + \frac{1}{N^v} \sum_{i=1}^{N^v} (BCE(\mathcal{S}_{i, mask}^{(t)}, \hat{\mathcal{S}}_{i, mask}^{(t)})
         + (p_{i}^{(t)} - \hat{p}_{i}^{(t)} )^2)\big)
        \label{eq:gn_error_cross_entropy}
    \end{aligned}
\end{equation}
for $n$ predictions, $N^v$ objects, $N^e$ edges and frame values in $[0,1]$ which is achieved by appending a sigmoid function at the end of the decoder (rounding values $\geq 0.5$ to 1 and values smaller to 0). We note that, despite providing RGB, depth and the object segmentation frame, we ask our models to only predict individual object segmentation masks. In our ablation study below, we evaluate the effects of removing some of these information such as position and velocity in nodes and edges.

\paragraph{Auto-Predictor Approach}
Contrary to the GN method, the AP approach does not consider explicitly defined edge attributes. Consequently, both approaches employ identical encoding and decoding phases (also in terms of network configurations) except that the AP architecture does not encode edges and nodes only contain visual observations, i.e. $v^{(t)} = (\mathcal{I}^{(t)}, \mathcal{S}_{v, mask}^{(t)}, \mathcal{D}^{(t)})$. In the processing phase, the GN architecture uses the full GN block to predict states of nodes, edges and global. In contrast, the AP architecture only predicts node states with a module we term \emph{physics predictor} and thus, constitutes a more lightweight block than the full GN block. As in the GN architecture, we use a new control input before each prediction. Thus, we update the latent node features with a new control $\bar{V}= \{v_{latent}^{(t)} \oplus \mathcal{C}^{(t)}_{latent} \mid  v_{latent}^{(t)} \in V(\hat{G}^{(t)}_{latent})\}$ with $\hat{G}^{(t)}_{latent} = (u,\bar{V},E)$ where $\oplus$ denotes concatenation. We point out that this concatenation distinguishes the AP from the O2P2 \cite{Janner2019} architecture. As in \cite{Janner2019}, the physics predictor contains the two functions $f_{interact}$ and $f_{trans}$ each implemented by an MLP. Altogether, the physics predictor defines an update rule for node states with $\hat{v}_{i, latent}^{(t+1)} = v_{i, latent}^{(t)} + f_{trans}(\bar{v}_{i, latent}^{(t)}) + \sum_{i \neq j} f_{interact}(\bar{v}_{i, latent}^{(t)}, \bar{v}_{j, latent}^{(t)})$ where $\bar{v}_{j, latent}^{(t)} \in \bar{V}$ denotes all remaining latent nodes. Further, all AP models are trained by optimizing the BCE term in Equation \ref{eq:gn_error_cross_entropy}. We further add a latent space loss term as suggested by \cite{Johnson2016} between the predicted latent node vectors and the ground truth latent vectors for which we trained a separate auto-encoder to memorize all samples of the \emph{3 objects} dataset (see Figure \ref{fig:gnmodel}). 

%While the full GN block predictions rely on edges, nodes, control commands and can deal with arbitrarily many edges, the physics predictor is more light-weight as its predictions rely on nodes, control commands and pairwise connections only. 

\paragraph{Baseline}
The baseline architecture is equivalent to the AP architecture but without $f_{trans}$ and $f_{interact}$. It uses $(\mathcal{I}^{(t)}, \mathcal{S}_{v, mask}^{(t)}, \mathcal{D}^{(t)})$ as input, conditions its prediction on $\mathcal{C}^{(t)}_{latent}$ and employs a CNN encoder, a transpose CNN decoder and an MLP for encoding controls (see supplementary material for details). We use identical hyperparameters as for the AP and GN (network sizes, learning rate, etc.) and train it by optimizing the BCE.

%For computing $n$ predictions, we repeatedly looped back into the input the previous output, i.e. we set $x_v^{(t+1)} = \hat{x}_v^{(t+1)}$ and replaced the encoding of $\mathcal{C}^{(t)}$ with an encoding of $\mathcal{C}^{(t+1)}$ to predict $\hat{x}_v^{(t+2)}$ and so on. 

\section{Experiments}
\begin{figure}[!ht]
    \centering
    \includegraphics[width=1.0\linewidth]{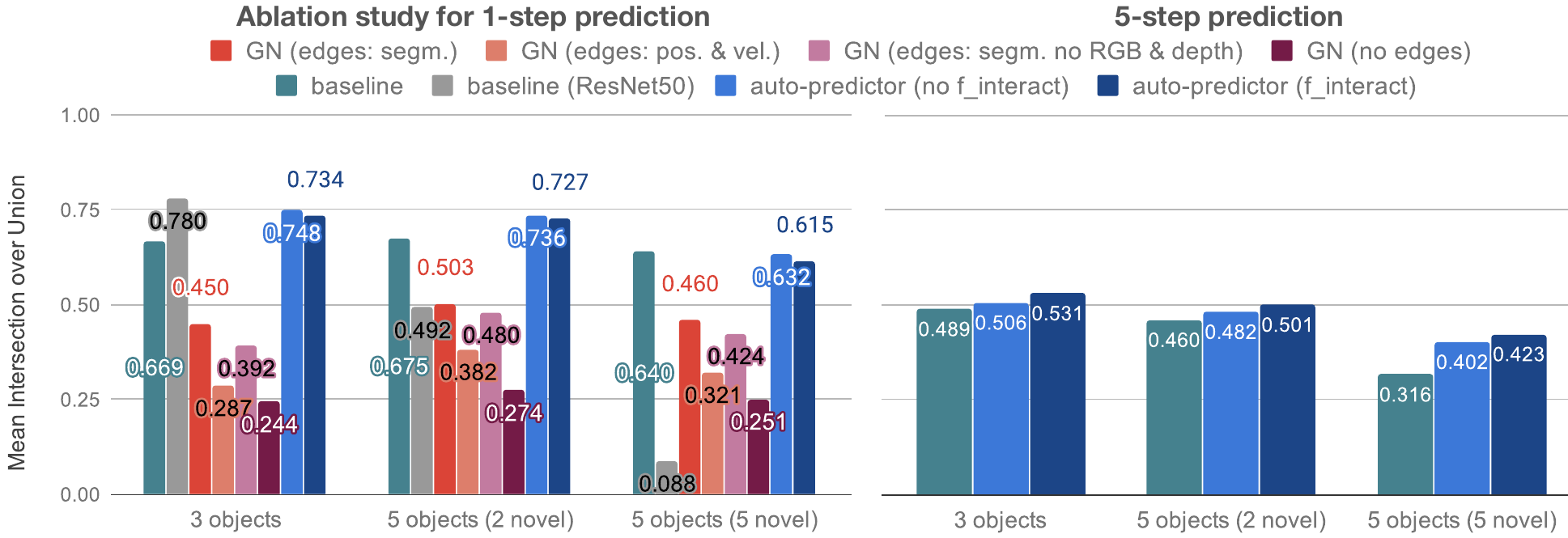}
    \caption{Our quantitative results for 1-step and 5-step prediction}
    \label{fig:quantresults}
\end{figure}

To assess the predictive capability of our models, we conducted quantitative and qualitative experiments. However, we restrict this work to the quantitative part due to space limitations (see our project website for all results). The quantitative experiments report the mean Intersection over Union (mean IoU) which measures the similarity between the ground truth object segment and the predicted segment averaged across all episodes, time steps and objects of a dataset. We use the following model abbreviations; \textbf{\emph{baseline (ResNet50)}}: the baseline with a ResNet50 encoder (no pre-training), \textbf{\emph{GN (edges: pos. \& vel.)}}: the GN model introduced above, \textbf{\emph{GN (edges: segm.)}}: the GN model based on visual information only, i.e. we remove all object positions/velocities in nodes and edges and use segmentation masks as edge attributes and replace the edge MLPs by a small CNN encoder/decoder (trained with BCE), \textbf{\emph{GN (edges: segm. no RGB \& depth.)}}: similar to \emph{GN (edges: segm.)} but RGB and depth is removed from the input, \textbf{\emph{GN (no edges)}}: the use of edges is deactivated, \textbf{\emph{auto-predictor (no f\_interact)}} the AP architecture without $f_{interact}$, i.e. no interactions are computed. In our ablation study in Figure \ref{fig:quantresults} (left) we depict performances of models trained and tested for 1-step predictions. It shows that both baseline and AP models outperform all edge attribute-based models by a significant margin which indicates a sensitivity with regard to the edge attribute choice. Figure \ref{fig:quantresults} (right) shows model performances trained and tested for 5-step prediction where we loop back the last prediction to the input for each new prediction. While baseline and the auto-predictor yield similar results for familiar objects, the auto-predictor is less prone to overfitting and generalizes better to novel shapes. Lastly, $f_{interact}$ seems to consistently improve the results across all datasets, indicating that letting the model to explicitly reason about object interactions is beneficial.
%not just that adding RGB and depth frames improve the performance only marginally \jean{over what? Adding this to what? Ground truth locations?} but more importantly 

\section{Discussion}
We presented our results for learning object-centric, visual dynamics models with two GNN approaches trained on occluded visual observations that involve complex objects and scene dynamics. The results indicate that the choice of edge attributes can significantly impact the predictive performance as the models not using edge attributes yield superior results over all edge attribute-based models. While the reason for this remains unknown to us, we conjecture an inferior selection of edge attributes can negatively impact the representation space of edge and node update functions in GN-based approaches. Overall, despite our results showing that vision-based motion prediction remains challenging, we see evidence for the benefits of imposing relational inductive biases in terms of generalization and advocate choosing lightweight approaches that implicitly reason about relations over ones that leave the specification of relations to human designers.

\newpage
% ------ uncomment for final submission ------
\subsubsection*{Acknowledgements}
The research leading to these results has received funding from the Toyota Research Institute (TRI), the European Union’s Horizon 2020 Research and Innovation programme under grant agreement No. 731761 (IMAGINE) and the International Center for Advanced Communication Technologies (InterACT). This article solely reflects the opinions and conclusions of its authors and not of TRI or any entity associated with Toyota.

%The research leading to these results has received funding from the International Center for Advanced Communication Technologies (InterACT).

\bibliographystyle{unsrtnat}
\bibliography{references}

\newpage

\section*{Supplementary Materials}
\subsection{Dataset Overview}
\begin{table}[h!]
\centering
\caption{Overview of used datasets in this work. All datasets without a specified number of training episodes denote they were only used for testing. Frames have a resolution of $160 \times 120$.}
\begin{tabular}{@{}lllll@{}}
\toprule
Dataset name  & \# training episodes & \# test ep. & \# novel object shapes & \# time step range \\ \midrule
3 objects     & 10 000              & 2 600           & -                     & 7...15             \\
5 objects (2 novel) & -                   & 1 690           & 2                     & 7...50             \\
5 objects (5 novel) & -                   & 1 520           & 5                     & 7...50             \\
2 large cubes & 1 658                & 415             & -                     & 7...50             \\ \bottomrule
\end{tabular}
\label{tab:data}
\end{table}

\subsection{Learning the Simulation Agent}
We learned the simulation agent with A3C \cite{Mnih2016} and a sparse reward function (+1 if the Euclidean distance between a randomly chosen object and all other objects was larger than 10 cm, 0 otherwise).  We used a continuous state space (represented by Euclidean centroid position and current velocity of objects) and discrete action space (left, right, forward, backward).

\subsection{Training the Models}
All models were trained for 13 epochs with Adam \cite{Kingma2014}, a learning rate of $0.001$ and a mini training batch size of 30. All layers of our networks used rectified linear unit (ReLU) activations followed by Batch Normalization \citep{Ioffe2015} layers except for the last layers of each network for which we used linear activations and no normalization. Note that we experimented with 10-20\% dropout in all layers except the last ones of each network but did not observe any benefits in training efficiency, qualitative results or measured performances (mean IoU). In the multi-step prediction experiments, we trained the AP and baseline models with a curriculum training strategy. Under this strategy, we divide the training of an $n$-step prediction model into $n$ stages and increase the task difficulty for the model gradually. First, we train the model for 1-step prediction for some time. Next, we use this model for training 2-step prediction. We repeat this procedure until we reach $n$-step prediction. Including all stages, we always train for a total of 13 epochs and divide training time uniformly among these stages.

\subsection{Neural Network Configurations}
\label{ap:nnconfigs}
We adopt the following syntax:
\begin{itemize}
    \item \emph{conv3x3-1-128}: convolutional layer with kernel size 3, stride 1 and 128 feature maps
    \item \emph{transpconv4x4-[1x2]-256}: transponse convolution layer with kernel size 4, stride 1x2 (x/y dimension) and 256 feature maps
    \item \emph{FC-32}: a fully-connected layer with 32 units
    \item \emph{maxpool}: a max pooling layer with pooling size 2 and stride 2 
\end{itemize}

\subsubsection*{Auto-Predictor}
\label{appendix:nnconfigs-ap}

\begin{table}[ht!]
\centering
\caption{Network configurations employed by the AP architecture.}
\begin{tabular}{cc}
\toprule
\multicolumn{2}{c}{\textbf{Auto-Predictor}} \\ \hline
\textbf{\begin{tabular}[c]{@{}c@{}}$f_{trans}$\\ MLP\end{tabular}} & \textbf{\begin{tabular}[c]{@{}c@{}}$f_{interact}$\\ MLP\end{tabular}} \\ \hline
FC-256 & FC-256 \\ \hline
FC-256 & FC-256 \\ \bottomrule
\end{tabular}
\label{tab:nnconfigs4}
\end{table}

\subsubsection*{Auto-Predictor and Graph Network}
\label{appendix:nnconfigs-apgn}

\begin{table}[H]
\centering
\caption{Network configurations employed by both the AP and the GN architecture.}
\begin{tabular}{ccll}
\toprule
\multicolumn{4}{c}{\textbf{Auto-Predictor / Graph Network}} \\ \hline
\multicolumn{1}{c}{\textbf{\begin{tabular}[c]{@{}c@{}}Node encoder\\ CNN\end{tabular}}} & \textbf{\begin{tabular}[c]{@{}c@{}}Node decoder \\ transpose CNN\end{tabular}} & \multicolumn{1}{c}{\textbf{\begin{tabular}[c]{@{}c@{}}Control / global\\ encoder MLP\end{tabular}}} & \multicolumn{1}{c}{\textbf{\begin{tabular}[c]{@{}c@{}}Control / global\\ encoder MLP (2)\end{tabular}}} \\ \hline
\multicolumn{1}{c}{\begin{tabular}[c]{@{}c@{}}conv3x3-1-128\\ conv3x3-1-128\end{tabular}} & transpconv2x2-1-256 & \multicolumn{1}{c}{FC-32} & \multicolumn{1}{c}{FC-6} \\ \hline
\multicolumn{1}{c}{maxpool} & transpconv2x2-2-256 & \multicolumn{1}{c}{FC-32} & \multicolumn{1}{c}{FC-6} \\ \hline
\multicolumn{1}{c}{\begin{tabular}[c]{@{}c@{}}conv3x3-1-256\\ conv3x3-1-256\end{tabular}} & transpconv4x4-[1x2]-256 & \multicolumn{1}{c}{FC-32} & \multicolumn{1}{c}{FC-6} \\ \hline
\multicolumn{1}{c}{maxpool} & transpconv3x2-2-256 & \multicolumn{1}{c}{} &  \\ \hline
\multicolumn{1}{c}{\begin{tabular}[c]{@{}c@{}}conv3x3-1-256\\ conv3x3-1-256\end{tabular}} & transpconv2x2-1-256 & \multicolumn{1}{c}{} &  \\ \hline
\multicolumn{1}{c}{maxpool} & transpconv2x2-1-128 &  &  \\ \hline
\multicolumn{1}{c}{\begin{tabular}[c]{@{}c@{}}conv3x3-1-256\\ conv3x3-1-256\end{tabular}} & transpconv2x2-2-128 &  &  \\ \hline
\multicolumn{1}{c}{maxpool} & transpconv3x3-1-128 &  &  \\ \hline
\multicolumn{1}{c}{\begin{tabular}[c]{@{}c@{}}conv3x3-2-256\\ conv3x3-2-256\end{tabular}} & transpconv3x3-2-128 &  &  \\ \hline
\multicolumn{1}{c}{maxpool} & transpconv3x3-1-128 &  &  \\ \hline
\multicolumn{1}{l}{} & transpconv3x3-2-128 &  &  \\ \hline
\multicolumn{1}{l}{} & transpconv3x3-1-2 &  &  \\ \bottomrule
\end{tabular}
\label{tab:nnconfigs1}
\end{table}

\subsubsection*{Graph Network}
\label{appendix:nnconfigs-gn}

\begin{table}[h!]
\centering
\caption{Network configurations employed by the GN architecture. We refer to the full GN block as \emph{Core}.}
\begin{tabular}{ccccc}
\toprule
\multicolumn{5}{c}{\textbf{Graph Network}} \\ \hline
\textbf{\begin{tabular}[c]{@{}c@{}}Edge encoder \\ MLP\end{tabular}} & \textbf{\begin{tabular}[c]{@{}c@{}}Edge decoder \\ MLP\end{tabular}} & \textbf{\begin{tabular}[c]{@{}c@{}}Core node \\ MLP\end{tabular}} & \textbf{\begin{tabular}[c]{@{}c@{}}Core edge\\ MLP\end{tabular}} & \textbf{\begin{tabular}[c]{@{}c@{}}Core global\\ MLP\end{tabular}} \\ \hline
FC-64 & FC-64 & FC-256 & FC-64 & FC-32 \\ \hline
FC-64 & FC-64 & FC-256 & FC-64 & FC-32 \\ \hline
FC-64 & FC-9 & FC-256 & FC-64 & FC-32 \\ \bottomrule
\end{tabular}
\label{tab:nnconfigs2}
\end{table}

\begin{table}[h!]
\centering
\caption{Network configurations employed by the GN architecture.}
\begin{tabular}{cccc}
\toprule
\multicolumn{4}{c}{\textbf{Graph Network}} \\ \hline
\multicolumn{1}{c}{\textbf{\begin{tabular}[c]{@{}c@{}}Position \& velocity \\ encoder MLP (nodes)\end{tabular}}} & \textbf{\begin{tabular}[c]{@{}c@{}}Position \& velocity \\ decoder MLP (nodes)\end{tabular}} & \textbf{\begin{tabular}[c]{@{}c@{}}Small edge \\ encoder CNN\end{tabular}} & \textbf{\begin{tabular}[c]{@{}c@{}}Small edge \\ decoder transpose CNN\end{tabular}} \\ \hline
\multicolumn{1}{c}{FC-32} & FC-32 & \begin{tabular}[c]{@{}c@{}}conv3x3-2-32\\ conv3x3-2-32\end{tabular} & \begin{tabular}[c]{@{}c@{}}transpconv2x2-1-64\\ transpconv2x2-2-64\end{tabular} \\ \hline
\multicolumn{1}{c}{FC-32} & FC-32 & conv3x3-2-16 & \begin{tabular}[c]{@{}c@{}}transpconv4x4-{[}1x2{]}-32\\ transpconv3x2-2-16\\ transpconv2x2-2-8\end{tabular} \\ \hline
 & FC-6 & conv3x3-2-5 & \begin{tabular}[c]{@{}c@{}}transpconv3x2-2-2\\ transpconv3x2-2-2\end{tabular} \\ \bottomrule
\end{tabular}
\label{tab:nnconfigs3}
\end{table}

\subsubsection*{Baseline}
The baseline encoder is similar to \emph{Node encoder CNN}, the decoder is similar to \emph{Node decoder transpose CNN} and the control encoder is similar to \emph{Control global encoder MLP}.

\end{document}